\documentclass{article}

\usepackage{eusipco}                      
\usepackage{amsfonts}
\usepackage{amsmath}
\usepackage{amssymb}

\title{Informed Source Separation:\\ A Bayesian Tutorial}

\name{Kevin H. Knuth\thanks{This work supported by the NASA SISM
Intelligent Systems Program and NASA ESTO AIST-QRS-04-3010.}}

\address{Intelligent Systems Division\\ NASA Ames Research Center\\
Moffett Field CA 94035 USA\\
\\
Knuth K.H. 2005. Informed source separation: A Bayesian tutorial. (Invited paper)\\ 
B. Sankur , E. Cetin, M. Tekalp , E. Kuruoglu (eds.), \\
Proceedings of the 13th European Signal Processing Conference (EUSIPCO 2005), Antalya, Turkey.}

\begin{document}

\maketitle

\begin{abstract}
Source separation problems are ubiquitous in the physical
sciences; any situation where signals are superimposed calls for
source separation to estimate the original signals. In this
tutorial I will discuss the Bayesian approach to the source
separation problem.  This approach has a specific advantage in
that it requires the designer to explicitly describe the signal
model in addition to any other information or assumptions that go
into the problem description. This leads naturally to the idea of
\emph{informed source separation}, where the algorithm design
incorporates relevant information about the specific problem. This
approach promises to enable researchers to design their own
high-quality algorithms that are specifically tailored to the
problem at hand.
\end{abstract}


\section{Understanding the Problem}

To gather information about the physical world, we deploy sensors
to make measurements and detect signals.  Our sensors, if properly
designed, will collect information about the signals of interest.
However, very often the signals of interest are comprised of a set
of discrete signals, which have been superimposed during
propagation, often with signals that are not of interest.  Thus
our sensors almost invariably detect a mixture of signals---some
interesting and some non-interesting. In more straightforward
applications, careful design of the sensors and application of
filters can limit the recordings to the signal of interest.
However, when this is not possible, more extreme steps need to be
taken.  This leads to a class of problems called source separation
problems.

There is no limit to the complications that may arise.
Superposition may be linear, or one of the infinite varieties of
nonlinear superposition.  If a set of sensors are used, there may
be time delays due to propagation of the signals from each source
to each detector, or there could be convolutions due to
reflections or differences in propagation speed through
intervening media coupled with diffraction.  To presume to be able
to construct a single algorithm that can deal with all of these
imaginable cases is unrealistic.  Instead, we must focus our
efforts on developing a methodology for designing robust
algorithms that are specific to the application at hand.  Only
then can we take advantage of the specific prior knowledge we
possess about each problem to increase our chances of reaching an
accurate and optimal solution. I call this approach \emph{informed
source separation}, which should be contrasted with blind source
separation, where very little is assumed to be known about the
problem.

The source separation problem can be viewed as an inference
problem, where one models a set of detected signals as a mixture
of a set of source signals.  It is important to remember that
inference is not deduction---it doesn't always work.  In difficult
problems, prior information goes a long way to help assure that we
reach an accurate solution.  This prior information can take many
forms, and can come into the problem at several different points.
I will show that this prior information can significantly
transform the source separation problem, and subsequently, the
algorithmic solution.

\section{Bayesian Probability Theory}

In this section, I give a brief description of the Bayesian
methodology.  I will focus on the use of probability theory to
describe our knowledge, and leave the details of the problem of
searching the hypothesis space for the optimal solution to other
authors. The crux of the methodology is Bayes' Theorem
\begin{equation}\label{eq:bayes}
p(model | data, I) = p(model|I) \frac{p(data | model, I)}{p(data |
I)}
\end{equation}
where $I$ represents our prior information.  The probability on
the left $p(model | data, I)$ is called the \emph{posterior
probability}.  It is the probability that a specific $model$
accurately describes the problem given the $data$ and our prior
information $I$. The first term on the right $p(model|I)$ is
called the \emph{prior probability}, or \emph{prior} for short.
The prior describes the degree to which we believe a specific
model is the correct description before we see any data, and thus
encodes our knowledge about the possible values of the model
parameters. The term in the numerator $p(data | model, I)$ is
called the \emph{likelihood}, which describes the degree to which
we believe that the model could have produced the observed data.
This term encodes both the process of making predictions with our
hypothesized model and the process of comparing these predictions
to our data, which is an important part of the scientific method.
The term in the denominator $p(data | I)$ is called the evidence.
In many parameter estimation problems, where we have a static
model and are merely estimating the values of its parameters, this
term simply acts a normalization factor. In problems where we are
testing one of a set of several models, this term becomes
extremely relevant as it can indicate the degree to which a model
is favored by the data.

The space of all considered models is called the hypothesis space.
Bayes' Theorem turns the source separation problem into a search
problem, where we search the hypothesis space for the most
probable model.  We can also look at Bayes' Theorem as a learning
rule since tells us how to update our prior knowledge when we
receive new data.  What we have learned from this data combined
with what we knew prior is described by the posterior probability.
In the next section I will show how the Bayesian method applied to
a well-defined set of prior information leads directly to Infomax
ICA.

\section{First Example: Infomax ICA}

In this section I will demonstrate how the Bayesian methodology
allows one to derive Infomax ICA \cite{Bell&Sejnowski}. While this
particular derivation has been published previously
\cite{Knuth:ICA99}, I present it here again in detail both to
assist with understanding the later derivations in this paper, and
also to clear up some common misconceptions surrounding ICA and
source separation. ICA is commonly considered to be a blind source
separation algorithm because we make a minimal number of
assumptions.  However, it is important point to note that no
algorithm is truly blind, and that the assumptions we make---even
if minimal in some sense---will have an affect on the performance
of an algorithm when applied to a given problem.

We begin by assuming that there are $N$ sources whose signals
propagate instantaneously to $N$ distinct detectors.  The signals
are assumed to superimpose linearly so that each detector records
a linear mixture of the source signals.  Furthermore, the entire
process is assumed to be noise-free. This leads to a simple
mathematical model that describes the recorded signals in terms of
the unknown sources\footnote{I have purposefully kept this in
component form so that it may be more easily compared with other
algorithms presented later in this tutorial.}
\begin{equation}\label{eq:infomax-model}
x_{it} = \sum_{j=1}^{N}{A_{ij} s_{jt}}
\end{equation}
where $x_{it}$ is the signal recorded at $i^{th}$ detector at time
$t$, $s_{jt}$ is the source signal emitted by the $j^{th}$ source
at time $t$, and $A_{ij}$ is the ``mixing matrix''.  The elements
of the mixing matrix serve to couple the sources to the detectors.
Physically each $A_{ij}$ describes how the signal propagates from
the source to the detector.  While the physical interpretation is
kept vague in the blind algorithm, we will see that the physical
interpretation is quite useful when deriving informed source
separation algorithms.

We can now apply Bayes' Theorem (\ref{eq:bayes}) to express the
probability of our model, $\mathbf{A}$ and $\mathbf{s}$, given our
data, $\mathbf{x}$
\begin{equation} \label{eq:bayes-bss}
p(\mathbf{A}, \mathbf{s} | \mathbf{x}, I) = p(\mathbf{A},
\mathbf{s} | I) \frac{p(\mathbf{x} | \mathbf{A}, \mathbf{s},
I)}{p(\mathbf{x} | I)},
\end{equation}
where $\mathbf{A}$ represents the entire matrix, $\mathbf{s}$
represents all of the source signals emitted by the sources, and
$\mathbf{x}$ represents all of our recorded data. Given the fact
that the source signals are independent of the propagation, we can
factor the prior probability $p(\mathbf{A}, \mathbf{s}|I)$ into
two terms $p(\mathbf{A}|I)$ and $p(\mathbf{s}|I)$.
\begin{equation}\label{eq:infomax-sym-posterior}
p(\mathbf{A}, \mathbf{s} | \mathbf{x}, I) = p(\mathbf{A} | I)
p(\mathbf{s} | I) \frac{p(\mathbf{x} | \mathbf{A}, \mathbf{s},
I)}{p(\mathbf{x} | I)}.
\end{equation}
Once we assign these probabilities, the problem is in some sense
solved. All we will need to do is to search all possible values of
the matrix $\mathbf{A}$ and the source waveshapes $\mathbf{s}$ to
find the case that is most probable. \footnote{In practice,
conducting this search is often the most difficult part of the
problem.} When we perform this search, the evidence in the
denominator will never contribute to the calculation since it
doesn't depend on the model parameters. So we can simplify the
problem further by writing (\ref{eq:infomax-sym-posterior}) as a
proportionality
\begin{equation}\label{eq:infomax-sym-prop}
p(\mathbf{A}, \mathbf{s} | \mathbf{x}, I) \propto p(\mathbf{A} |
I) p(\mathbf{s} | I) p(\mathbf{x} | \mathbf{A}, \mathbf{s}, I).
\end{equation}

Now with the basic model in hand, we can construct a likelihood
function.  In this case, we are assuming that the recording
process is noise-free; thus we will assign a delta-function
likelihood function for each datum point
\begin{equation}\label{infomax-single-likelihood}
p(x_{it} | \mathbf{A}, \mathbf{s}_t, I) = \delta \bigl(x_{it} -
\sum_{j=1}^{N}{A_{ij} s_{jt}} \bigr),
\end{equation}
where I have used $\mathbf{s}_t$ to represent all of the $N$
source amplitudes emitted at time $t$.  This delta function
likelihood states, very strongly, that we believe that our model
of the source separation problem (\ref{eq:infomax-model}) is
correct. The recordings are independent of one other, so the
likelihood function for our entire data set is merely the product
of likelihoods (\ref{infomax-single-likelihood}) for each detector
and each time step
\begin{equation}\label{infomax-joint-likelihood}
p(\mathbf{x} | \mathbf{A}, \mathbf{s}, I) =
\prod_{i=1}^{N}{\prod_{t=1}^{T}{\delta \bigl(x_{it} -
\sum_{j=1}^{N}{A_{ij} s_{jt}} \bigr)}},
\end{equation}
where $T$ is the number of time steps.

Next we assume that the probability density of the amplitude of
the individual source signals has a positive kurtosis (also known
as leptokurtotic or super-Gaussian). With this assumption, we can
assign a prior probability for the amplitudes of the signals
emitted by the sources.  Without such a prior, this problem has an
infinite number of perfectly good solutions.  This prior
information will serve to make the problem soluble in cases where
it is correct, while risking the possibility of incorrect
solutions in cases where this assumption does not hold.  We will
write
\begin{equation}
p(s_{jt} | I) = q_j(s_{jt}),
\end{equation}
where $q_j(s_{jt})$ is the probability that the $j^{th}$ source
could have a given amplitude at any time $t$.  We could easily
follow Bell \& Sejnowski \cite{Bell&Sejnowski} and assign the
derivative of a sigmoid function, which is a leptokurtotic density
function. However, for the purposes of generalization, we will
just write it as $q_j$. Assuming that the sources are independent
of one another, we have
\begin{equation}\label{eq:infomax-prior-s}
p(\mathbf{s} | I) = \prod_{j=1}^{N}{\prod_{t=1}^{T}{q_i(s_{jt})}}.
\end{equation}

We now assume that we know nothing about the mixing matrix. We
will encode this knowledge by assigning a uniform
prior\footnote{The astute reader will recognize that each matrix
element acts as a scaling parameter in the problem. For this
reason, a more accurate noninformative prior would be the
appropriate Jeffrey's prior for the matrix $\mathbf{A}$.} for the
value of any given matrix element $A_{ij}$ as long as it is within
a ``reasonable'' range
\begin{equation}
p(A_{ij} | I) =
   \left\{ \begin{array}{rl}
       c & \mbox{if}~~A_{min} \geq A_{ij} \geq A_{max} \\
       0 & \mbox{if}~~A_{ij} < A_{min},~~A_{max} > A_{ij}\end{array} \right.
\end{equation}
where $c = (A_{max}-A_{min})^{-1}$. For the entire mixing matrix,
we can assign a uniform joint prior
\begin{align}\label{eq:infomax-prior-A}
p(\mathbf{A} | I) & = \prod_{i=1}^{N}{\prod_{j=1}^{N}{p(A_{ij} |
I)}}\\
& = \{ \begin{array}{rl}
       C & \mbox{if}~~\forall A_{ij},~~A_{min} \geq A_{ij} \geq A_{max} \\
       0 & \mbox{if}~~\exists A_{ij},~\mbox{s.t.}~~A_{ij} < A_{min},~~A_{max} > A_{ij}\end{array}  \nonumber
\end{align}
where $C = cN^2$. With our likelihood and priors all defined, we
are now ready to re-write the posterior probability
(\ref{eq:infomax-sym-prop}) and begin searching for the most
probable parameter values.

However, this search will not be easy.  Much of the effort in
Bayesian inference is to limit the number of parameters to search
over, or to come up with a clever heuristic to perform the search.
Furthermore, it is often easier to work with the logarithm of the
posterior (\ref{eq:infomax-sym-posterior}). This neatly separates
the posterior into a sum of the log priors plus the log
likelihood.

We will begin by reducing the number of parameters of interest,
and will conclude by taking the logarithm.  First we reason that
if we knew the mixing matrix $\mathbf{A}$, or better yet, its
inverse $\mathbf{A}^{-1}$, we could apply it to the data to
recover an estimate of the source signals. Surely this is not
ideal, but it is easier than searching the entire multidimensional
parameter space, which would be $N^2+NT$ parameters. To do this we
use the fact that probabilities sum to one, and \emph{marginalize}
over all possible values of the source signal amplitudes, written
symbolically as
\begin{align}\label{eq:infomax-marginalized-symbolic-posterior}
p(\mathbf{A} | \mathbf{x}, I) & = \int{d\mathbf{s}~~ p(\mathbf{A},
\mathbf{s} | \mathbf{x}, I)}\\
& \propto \int{d\mathbf{s}~~ p(\mathbf{A} | I) p(\mathbf{s} | I)
p(\mathbf{x} | \mathbf{A}, \mathbf{s}, I)}\nonumber\\  & \propto
p(\mathbf{A} | I) \int{d\mathbf{s}~~p(\mathbf{s} | I) p(\mathbf{x}
| \mathbf{A}, \mathbf{s}, I)} \nonumber
\end{align}
where the integral sign represents all $NT$ integrals over each of
the $s_{jt}$. Substituting our likelihood
(\ref{infomax-joint-likelihood}), and source density prior
(\ref{eq:infomax-prior-s}), we have $NT$ integrals to solve
\begin{equation}
\int{ds_{11} \cdots
\int{ds_{NT}~~\prod_{j=1}^{N}{\prod_{t=1}^{T}{q_j(s_{jt}) \delta
\bigl(x_{it} - \sum_{j=1}^{N}{A_{ij} s_{jt}}\bigr)}}}}.
\end{equation}
The delta functions easily allow us to solve each of the integrals
simply by introducing a change of variables where $w_{it} = x_{it}
- \sum_{j=1}^{N}{A_{ij} s_{jt}}$. We then have that $\det
\mathbf{A}~d\mathbf{s} = d\mathbf{w}$, and that $s_{jt} =
\sum_{i=1}^{N}{A_{ij}^{-1} (x_{it} - w_{it})}$, so that the
integral becomes
\begin{equation}
\frac{1}{\det A} \int{\cdots \int{dw_{NT}
\prod_{j=1}^{N}{\prod_{t=1}^{T}{q_j (\sum_{i=1}^{N}{A_{ij}^{-1}
(x_{it} - w_{it})}) \delta(w_{it})}}}}.
\end{equation}
The delta functions now all select $w_{it} = 0$, and we have as a
result
\begin{equation}
\frac{1}{\det A} \prod_{j=1}^{N}{\prod_{t=1}^{T}{q_j
\biggl(\sum_{i=1}^{N}{A_{ij}^{-1} x_{it}} \biggr)}}.
\end{equation}
Substituting this result into
(\ref{eq:infomax-marginalized-symbolic-posterior}) we get
\begin{equation}
p(\mathbf{A} | \mathbf{x}, I) \propto p(\mathbf{A} | I)
\frac{1}{\det A} \prod_{j=1}^{N}{\prod_{t=1}^{T}{q_j
\biggl(\sum_{i=1}^{N}{A_{ij}^{-1} x_{it}} \biggr)}},
\end{equation}
which when taking the logarithm gives
\begin{multline}\label{eq:log-bss-posterior}
\log p(\mathbf{A} | \mathbf{x}, I) = K +\\ \log p(\mathbf{A} | I)
- \log \det A + \sum_{j=1}^{N}{\sum_{t=1}^{T}{\log q_j
\biggl(\sum_{i=1}^{N}{A_{ij}^{-1} x_{it}} \biggr)}},
\end{multline}
where $K$ is the logarithm of the constant implicit in the
proportionality. By varying $\mathbf{A}$ to maximize the log
posterior above, we can solve for the optimal mixing matrix.  The
way this is done in ICA is to take the derivative with respect to
the inverse of $\mathbf{A}$ and to use this in a gradient ascent
learning rule.  Specifically, if we assign the mixing matrix prior
according to (\ref{eq:infomax-prior-A}), and write $W_{ij} =
A_{ij}^{-1}$, we get the familiar Infomax ICA gradient ascent
learning rule \cite{Bell&Sejnowski, Knuth:ICA99}
\begin{align}\label{eq:infomax-learning-rule}
\nonumber \Delta W_{ij} & = \frac{\partial}{\partial
W_{ij}}\biggl[ - \log \det A + \sum_{j=1}^{N}{\sum_{t=1}^{T}{\log
q_j \biggl(\sum_{i=1}^{N}{A_{ij}^{-1} x_{it}} \biggr)}} \biggr]\\
& = A_{ji} + \sum_{j=1}^{N}{\sum_{t=1}^{T}{x_{jt}
\biggl(\frac{q_i'(u_{it})}{q_i(u_{it})} \biggr)_j}},
\end{align}
where $u_{it} = \sum_{i=1}^{N}{A_{ij}^{-1} x_{it}}$.

However, strictly speaking, this rule doesn't lead to the optimal
separation matrix $\mathbf{W}$, since the maximum value of the
posterior with respect to variations of $\mathbf{A}$ will not be
identical to the maximum value of the posterior with respect to
$\mathbf{A^{-1}}$. This is due to the fact that probability
densities transform differently than functions. Since
\begin{equation}\label{eq:jacobian}
p(\mathbf{A^{-1}}|\mathbf{x},I) = p(\mathbf{A}|\mathbf{x},I)
\biggl| \frac{\partial \mathbf{A^{-1}}}{\partial \mathbf{A}}
\biggr|^{-1},
\end{equation}
if we define
\begin{align}
\mathbf{\hat{A}} & = \underset{\mathbf{A}}{\textrm{arg
max}}~~p(\mathbf{A}|\mathbf{x},I)\\
\mathbf{\breve{W}} & =
\underset{\mathbf{A^{-1}}}{\textrm{arg
max}}~~p(\mathbf{A}|\mathbf{x},I)\label{eq:infomax-W-breve} \\
\mathbf{\hat{W}} & = \underset{\mathbf{A^{-1}}}{\textrm{arg max}}
~~p(\mathbf{A^{-1}}|\mathbf{x},I)\label{eq:correct-W-hat}
\end{align}
in general, we have that
\begin{equation}
\mathbf{\breve{W}} \neq
\mathbf{\hat{W}}~~\mbox{and}~~\mathbf{\hat{W}} \neq
\mathbf{\hat{A}}^{-1},
\end{equation}
where $\mathbf{\hat{W}}$ is the optimal separation matrix,
$\mathbf{\hat{A}}$ is the optimal mixing matrix, and
$\mathbf{\breve{W}}$ is the Infomax ICA solution. Thus the inverse
of the optimal estimate of the mixing matrix does not equal the
optimal estimate of the separation matrix.\footnote{The classic
example of this is the fact that the frequency at which a given
blackbody spectrum has maximum energy density is different than
the wavelength at which it has maximum energy density.} However,
the ICA solution is actually neither of these. If we were really
interested in finding the most probable inverse of the mixing
matrix, we should have used (\ref{eq:jacobian}) to write the
posterior for $\mathbf{A^{-1}}$ and solved for its most probable
value (\ref{eq:correct-W-hat}). As a result the standard technique
(\ref{eq:infomax-learning-rule}) and (\ref{eq:infomax-W-breve})
leads to a biased separation.

That being said, there is much one can learn from this derivation
of the Infomax ICA algorithm.  Certainly one obtains the same
answer when deriving it from the information-theoretic viewpoint;
this being due to the duality between probability theory and
information theory \cite{Knuth:2005}.  However, the Bayesian
derivation has several distinct advantages.  First, all of the
assumptions that go into the algorithm are made explicit.  We see
that the sigmoidal nonlinearity in the original derivation is
merely related to the derivative of the prior probability for the
source amplitude density.  This answers one of the common
questions that arises: Why does ICA have problems separating pure
sinusoids?  The answer is clear; sinusoids have bimodal amplitude
histograms, which is a severe deviation from our prior expectation
of the super-Gaussian prior probability that we have assigned
explicitly, and Bell \& Sejnowksi assigned implicitly
\cite{Knuth:ME97}. Modifications to this prior to allow for
sub-Gaussian densities typically do not improve the situation
mainly because they are essentially smoothed uniform densities,
which are non-informative. If you want to separate sinusoids, you
need to include this relevant information in the design of the
algorithm.

Second, why does ICA assume the same number of sources as
detectors?  In this derivation one can see that the integral is
not analytically solvable if we do not make such an assumption. In
addition, if we would have assumed that the recorded signals were
noisy and assigned a Gaussian likelihood, we would again not be
able to perform the integration analytically past the first
integral. The noise-free square mixing matrix allows for an
analytic marginalization over the source waveshapes resulting in a
straightforward and elegant solution. However, this elegant
solution will break when pushed too far.

Last, another common question that arises is: Are
Gaussian-distributed signals separable? Often the answer is
`yes'---you just have to rely on additional or different prior
information. This is why understanding the information that goes
into the design of an algorithm is so important. It allows you to
better understand the range of applicability of an algorithm and
how to fix it when it doesn't work.  This is why I prefer the
Bayesian approach to source separation.  It requires you to make
all of this explicit.

\section{Incorporating Prior Knowledge}

In this section I will demonstrate another advantage to the
Bayesian approach.  We will modify the algorithm to account for a
simple piece of prior information.  Let's say that we know that
the speeds of propagation of the signals remain instantaneous, but
that the signals follow an inverse-square propagation law. Such
knowledge implies that the coefficients of the mixing matrix are
dependent on the relative positions of the sources and detectors.
How can we use this information if we have no knowledge of the
source-detector distances?

First, if we did know the distance $r_{ij}$ from source $j$ to
detector $i$, the mixing matrix element $A_{ij}$ would follow the
inverse-square propagation law
\begin{equation}
A_{ij} = \frac{1}{4 \pi r_{ij}^2}.
\end{equation}
However, we may know that the source must be within some maximum
distance $R$ from the detector.  If it could be anywhere in the
three-dimensional space surrounding the detector, we can assign a
uniform probability for its position $(r,\theta,\phi)$ within any
volume element of that space
\begin{equation}
p(r,\theta,\phi | I) = \frac{1}{V} = \frac{3}{4 \pi R^3},
\end{equation}
where $V$ is the spherical volume of radius $R$ surrounding the
detector.  This is the prior probability that the source is at any
position with respect to the detector.  However, we only need the
probability that the source is some distance $r$ from the
detector.  We obtain this by marginalizing over all possible
values of the angular coordinates
\begin{align}
p(r | I) & = \int_{0}^{2\pi}{d\phi~~\int_{0}^{\pi}{\sin \theta
d\theta~r^2 p(r,\theta,\phi | I)}}\\
& = \int_{0}^{2\pi}{d\phi~~\int_{0}^{\pi}{\sin \theta d\theta~r^2
\frac{3}{4 \pi R^3}}}\\\nonumber & = \frac{3r^2}{R^3}.
\end{align}
The prior on the source-detector distance is very reassuring since
it is naturally invariant with respect to coordinate rescaling
(change of variables).  Specifically if we introduce a new
coordinate system so that $\rho = a r$ and $P = a R$ with $a > 0$,
we find equating the probabilities around $\rho+d\rho$ and $r+dr$
that
\begin{align}
\bigl|p(\rho | I) d\rho \bigr| & = \bigl|p(r | I) dr \bigr|\\
\nonumber p(\rho | I)~\bigl|d\rho \bigr| & = p(r | I)~\bigl|dr \bigr|\\
\nonumber p(\rho | I) & = p(r | I)~\bigl|\frac{d\rho}{dr} \bigr|^{-1}\\
\nonumber p(\rho | I) & = \frac{3r^2}{R^3}~\bigl|\frac{d\rho}{dr}
\bigr|^{-1}\\
\nonumber p(\rho | I) & =
\frac{3a\rho^2}{P^3}~\bigl|a\bigr|^{-1}\\
\nonumber p(\rho | I) & = \frac{3\rho^2}{P^3}.
\end{align}
Since this prior is invariant with respect to coordinate
rescaling, we can measure distances using any units we wish.

We can now use this to derive a prior for the mixing matrix
element.  First write the joint probability using the product rule
\begin{equation}
p(A_{ij},r | I) = p(r|I) p(A_{ij} | r, I).
\end{equation}
The first term on the right is the source-detector distance prior,
and the second term is a delta function described by the hard
constraint of the inverse-square law. \footnote{Some readers may
wonder why I go through the difficulty of using delta functions
rather than computing the Jacobians and just performing a change
of variables with the probability densities as we did before when
demonstrating invariance with respect to rescaling.  The reason is
that in more complex problems where the parameter of interest
depends on multiple other parameters, the change of variables
technique becomes extremely difficult.  Care must be taken when
using delta functions, however, since the argument needs to be
written so that it is solved for the parameter of interest, in
this case $A_{ij}$ rather than another parameter such as $r$.}
These assignments give
\begin{equation}
p(A_{ij},r | I) = \frac{3r^2}{R^3}~\delta \bigl(A_{ij} - (4 \pi
r^2)^{-1} \bigr).
\end{equation}
We now marginalize over all possible values of $r$
\begin{equation}
p(A_{ij} | I) = \int_{0}^{R}{dr~~\frac{3r^2}{R^3}~\delta
\bigl(A_{ij} - (4 \pi r^2)^{-1} \bigr)}.
\end{equation}
To do this we will need to make a change of variables again by
defining $u = (A_{ij} - (4 \pi r^2)^{-1})$, so that
\begin{align}
r^2 & = [(4 \pi)(A_{ij}-u)]^{-1}\\
r^3 &= [(4 \pi)(A_{ij}-u)]^{-3/2}
\end{align}
and $du = (2 \pi r^3)^{-1} dr$, which can be rewritten as
\begin{equation}
dr = 2^{-3/2} (2\pi)^{-1/2} (A_{ij}-u)^{-3/2} du
\end{equation}
giving us
\begin{equation}
p(A_{ij} | I) = 2^{-4} \pi^{-3/2} \frac{3}{R^3}
\int_{-\infty}^{u_{max}}{du~~ (A_{ij}-u)^{-5/2} \delta(u)},
\end{equation}
where $u_{max} = A_{ij}-(4\pi R^2)^{-1}$. The delta function will
select $u=0$ as long as it is true that $u_{max}>0$ or
equivalently that $r<R$. If this is not true, the integral will be
zero. If this hard constraint of the sources being within a
distance $R$ of the detectors causes problems, your choice for $R$
was wrong.  The result is
\begin{equation}
p(A_{ij} | I) = \frac{3}{16 \pi^{3/2} R^3} A_{ij}^{-5/2}.
\end{equation}
so that the prior for the mixing matrix elements is proportional
to $A_{ij}^{-5/2}$.  Readers more familiar with statistics may
note (and perhaps worry) that this prior is improper since it
blows up as $A_{ij}$ goes to infinity.  This is not a practical
concern as long as the sources are not allowed to got off to
infinity.  Other readers may note that this prior depends
explicitly on the value of the maximum source-detector distance
$R$. More accurate knowledge about the value of $R$ will lead
naturally to a more appropriate prior probability, resulting in
more accurate source separation results. Once the value of $R$ has
been chosen, this prior can be inserted into
(\ref{eq:log-bss-posterior}) to generate a source separation
algorithm that accounts for this prior knowledge.

As one can imagine, these algorithms can be made arbitrarily more
detailed depending on the prior information available. If one has
information about the absolute positions of the detectors, such as
in a sensor web, and probable locations of the sources, one can
derive more accurate prior probabilities for the source-detector
distances \cite{Knuth:SPIE98} \footnote{The author would like to
thank Vivek Nigam for pointing out errors in the SPIE98 paper,
which will be corrected in a future version available at
http://www.arxiv.org/abs/physics/0205069}. This leads naturally to
more accurate prior probabilities for the mixing matrix elements,
which in turn lead to better results.

In an ICA-style gradient ascent learning rule
(\ref{eq:log-bss-posterior}, \ref{eq:infomax-learning-rule}) these
mixing matrix priors act as an additive term biasing the update
rule toward the solutions suggested by the prior knowledge. From
this perspective, the prior can be viewed as a regularizer.
However, one shouldn't get too carried away with this viewpoint
since rather than being devised in an \textit{ad hoc} manner,
these priors can be carefully designed based on the specific prior
information possessed by the algorithm designer. It would have
been very difficult to guess the prior we have just derived above.

\section{Separation and Localization}

Now that we have introduced the idea of the relative positions of
the sources and detectors, we can take this problem even further
and attempt not only to separate the sources, but also to localize
them. Here I present results from an earlier paper where we
considered the relationship between source separation and source
localization \cite{Knuth&Vaughan}.

We consider the problem of neural source estimation in
electroencephalography (EEG) where we have multiple neural sources
in the brain and multiple recording electrodes.  Each source $j$
emitting a signal $s_j$ will have some three-dimensional position
in the brain $\mathbf{p}_j$. In addition, these sources are such
that they often emit dipolar current fields, so we must also be
concerned about their orientation $\mathbf{q}_j$.  The mixing
matrix $\mathbf{A}$ again describes the coupling between the
sources and the detectors.  In the case of electrophysiology,
often much is known about this coupling since the electrodynamics
of current flow through tissue is well-understood and can be
modelled in detail using magnetic resonance imaging (MRI) derived
head models. In this problem, the electric currents propagate
nearly instantaneously throughout the head and superimpose
linearly resulting in signals $\mathbf{x}$ recorded by the
detectors. Using Bayes' Theorem, we can write the posterior
symbolically as
\begin{equation}
p(\mathbf{p}, \mathbf{q}, \mathbf{A}, \mathbf{s} | \mathbf{x}, I)
\propto p(\mathbf{p}, \mathbf{q}, \mathbf{A}, \mathbf{s} | I)
p(\mathbf{x}| \mathbf{p}, \mathbf{q}, \mathbf{A}, \mathbf{s}, I),
\end{equation}
where $\mathbf{p}$ represents the positions of all the sources in
the model, and similarly for the other non-subscripted parameters.
The model we have chosen is redundant in the sense that the mixing
matrix depends on the relative positions and orientations of the
sources to the detectors. This allows us to simplify some of the
terms above, and factor the prior using the product rule
\begin{multline}
p(\mathbf{p}, \mathbf{q}, \mathbf{A}, \mathbf{s} | \mathbf{x}, I)
\propto p(\mathbf{x}|
\mathbf{A}, \mathbf{s}, I) \times\\
p(\mathbf{A} | \mathbf{p}, \mathbf{q}, I) p(\mathbf{p} | I)
p(\mathbf{q}| \mathbf{p}, I) p(\mathbf{s} | I).
\end{multline}
These priors show that some model parameters are dependent on
others, such as the prior for the mixing matrix.  The orientations
of the sources depend on their position in the brain since the
orientations are determined by the histology of that particular
neural source.

Now if we assume that we know nothing about the source positions,
nor how they affect the mixing matrix, the prior $p(\mathbf{A} |
\mathbf{p}, \mathbf{q}, I)$ reduces to $p(\mathbf{A} | I)$.
Marginalizing over all source positions and orientations using
uniform priors, we recover the basic source separation problem
(\ref{eq:infomax-sym-posterior})
\begin{equation}
p(\mathbf{A}, \mathbf{s} | \mathbf{x}, I) \propto p(\mathbf{x}|
\mathbf{A}, \mathbf{s}, I) p(\mathbf{A} | I) p(\mathbf{s} | I).
\end{equation}
With the appropriate probability assignments, we could recover the
Infomax ICA algorithm, or perhaps another source separation
algorithm that is better suited for the job.

However, let's see what happens if we change our focus and
concentrate on the source positions rather than the mixing matrix.
We will describe the propagation of the signals from the sources
to the detectors with a \emph{forward model} that takes into
account the electrodynamics of the physical situation. For
simplicity, we will write this symbolically as a function
\begin{equation}
A_{ij} = F(\mathbf{d}_i, \mathbf{p}_j, \mathbf{q}_j),
\end{equation}
where $\mathbf{d}_i$ is the position of the $i^{th}$ detector,
$\mathbf{p}_j$ and $\mathbf{q}_j$ are the position and orientation
of the $j^{th}$ source. This function will play a role in our
prior probability for $A_{ij}$
\begin{equation}\label{eq:ese-prior-A}
p(\mathbf{A} | \mathbf{p}, \mathbf{q}, I) =
\prod_{i=1}^{M}{\prod_{j=1}^{N}{\delta \bigl(A_{ij} -
F(\mathbf{d}_i, \mathbf{p}_j, \mathbf{q}_j)\bigr)}},
\end{equation}
where we are assuming that there are $N$ sources and $M$
detectors.

Next we will assign a Gaussian likelihood to encode that we have
uncertainties about how well the model describes the experimental
data. The idea is to assume we know the expected squared deviation
$\sigma^2$ between the predicted and observed results.  Given that
we know this expected squared deviation, the principle of maximum
entropy \cite{Jaynes:book} says that the Gaussian distribution is
the most honest quantification of this knowledge.
\begin{equation}\label{eq:ese-likelihood}
p(\mathbf{x}| \mathbf{A}, \mathbf{s}, I) =
\prod_{i=1}^{M}{\prod_{t=1}^{T}{\exp \biggl[ -\frac{(x_{it} -
\sum_{l=1}^{N}{A_{il}s_{lt}})^2}{2\sigma_i^2} \biggr]}}.
\end{equation}
It is important to note that this does not imply that we believe
the noise is Gaussian distributed, it merely implies that we know
something about the expected squared deviation.\footnote{Jaynes
has an excellent chapter where he works through this common
misconception \cite{Jaynes:book} ch. 7, specifically 7.7.} If in
fact we do not know the actual value of $\sigma$, we can always
marginalize over it to obtain a more conservative probability
density related to Student's t distribution.

We now marginalize the posterior to get rid of the nuisance
parameters $A_{ij}$
\begin{multline}
p(\mathbf{p}, \mathbf{q}, \mathbf{s} | \mathbf{x}, I) \propto
p(\mathbf{p} | I) p(\mathbf{q}| \mathbf{p}, I) p(\mathbf{s} | I)\\
\int{d\mathbf{A}~~p(\mathbf{x}| \mathbf{A}, \mathbf{s}, I)
p(\mathbf{A} | \mathbf{p}, \mathbf{q}, I)}.
\end{multline}
With our probability assignments (\ref{eq:ese-likelihood}) and
(\ref{eq:ese-prior-A}), and writing $F_{ij} = F(\mathbf{d}_i,
\mathbf{p}_j, \mathbf{q}_j)$ the integrals become
\begin{multline}
\int d\mathbf{A}_{ij}~~\prod_{i=1}^{M} \prod_{t=1}^{T}{\exp
\biggl[ -\frac{(x_{it}
- \sum_{l=1}^{N}{A_{il}s_{lt}})^2}{2\sigma_i^2} \biggr]} \times\\
\prod_{j=1}^{N}{\delta \bigl(A_{ij} - F_{ij}\bigr)},
\end{multline}
which gives
\begin{equation}
\prod_{i=1}^{M}{\prod_{t=1}^{T}{\exp \biggl[ -\frac{(x_{it} -
\sum_{l=1}^{N}{F_{il}s_{lt}})^2}{2\sigma_i^2} \biggr]}}.
\end{equation}
Simplifying the notation further by writing $\hat{x}_{it} =
\sum_{l=1}^{N}{F_{il}s_{lt}}$,\footnote{Note that
$\hat{\mathbf{x}}$ are the predicted recordings based on the
sources $\mathbf{s}$ and the forward model.} the marginalized
posterior is then
\begin{multline}
p(\mathbf{p}, \mathbf{q}, \mathbf{s} | \mathbf{x}, I) \propto
p(\mathbf{p} | I) p(\mathbf{q}| \mathbf{p}, I) p(\mathbf{s} | I) \times\\
\prod_{i=1}^{M}{\prod_{t=1}^{T}{\exp \biggl[ -\frac{(x_{it} -
\hat{x}_{it})^2}{2\sigma_i^2} \biggr]}}.
\end{multline}

The remaining priors provide the potential for the introduction of
a significant amount of prior information.  In this demonstration
however, I will simply assign uniform priors.  Taking the
logarithm of the posterior results in
\begin{equation}
\log p(\mathbf{p}, \mathbf{q}, \mathbf{s} | \mathbf{x}, I) =
-\sum_{i=1}^{M}{\sum_{t=1}^{T}{\frac{(x_{it} -
\hat{x}_{it})^2}{2\sigma_i^2}}} + C,
\end{equation}
where $C$ is the logarithm of the implicit proportionality
constant. Maximizing this log posterior results in minimizing the
familiar chi-squared `cost function'
\begin{equation}
\chi^2 = \sum_{i=1}^{M}{\sum_{t=1}^{T}{\frac{(x_{it} -
\hat{x}_{it})^2}{2\sigma_i^2}}} + C,
\end{equation}
which is a common procedure in electromagnetic source
localization.

From this example we see that based on the parameters of interest
and the prior information we include, a source separation problem
can become a source localization problem. The lesson here is that
the Bayesian formalism is the structure that underlies not only
source separation and source localization problems, but rather
signal processing in general. In fact, many familiar
techniques---even the Fourier transform
\cite{Bretthorst:1988}---have their basis in the Bayesian
methodology and can be significantly improved by understanding the
underlying models and assumptions that go into the algorithm.

\section{Beyond Separation}

We can take these ideas further by developing signal models that
include parameters that allow us to characterize or describe the
signals in different ways.  In this example, we again consider EEG
signals. Typically an experimenter will design an experiment and
record data from multiple experimental trials.  The standard
analysis technique consists of averaging the data across trials to
reduce the effects of noise (`noise' meaning signals that are
either not understood or not interesting).  The implicit signal
model that is employed is the signal plus noise (SPN) model where
we assume that we have a single stereotypic source waveshape
$s(t)$ that is produced every trial in addition to ongoing noise
$\eta(t)$.\footnote{I have changed notation here slightly where I
am now writing these signals as functions of time.  This notation
is more clear later when we are required to describe latency
shifts of the neural response.} The data that is recorded in an
electrode can be modelled as
\begin{equation}
x_{r}(t) = s(t) + \eta_{r}(t),
\end{equation}
where $r$ indexes one of the $R$ trials and $t$ indexes the
measurements at $T$ discrete time points. \footnote{Note that the
Bayesian methodology does not require that these measurements be
equally spaced in time.  This is a distinct advantage when dealing
with `missing data' problems.} Using Bayes' Theorem we have,
\begin{equation}
p(\mathbf{s} | \mathbf{x}, I) \propto p(\mathbf{x} | \mathbf{s},
I) p(\mathbf{s} | I).
\end{equation}
Relying on arguments laid out in the previous section, I will
assign a Gaussian likelihood, and a uniform prior for $s$.  The
log posterior is then
\begin{equation}
\log p(\mathbf{s} | \mathbf{x}, I) =
-\sum_{r=1}^{R}{\sum_{t=1}^{T}{\frac{(x_{r}(t) -
s(t))^2}{2\sigma^2}}} + C.
\end{equation}
We can find the maximum of the log posterior by taking the
derivative with respect to each $s(t)$ and setting it equal to
zero. For a particular time $t'$ we have
\begin{align}
\frac{\partial}{\partial s(t')}\log p(\mathbf{s} | \mathbf{x}, I)
&= \frac{\partial}{\partial s(t')} \biggl(
-\sum_{r=1}^{R}{\sum_{t=1}^{T}{\frac{(x_{r}(t) -
s(t))^2}{2\sigma^2}}} +
C \biggr)\nonumber\\
&= -\frac{1}{2\sigma^2}\sum_{r=1}^{R}{\frac{\partial}{\partial
s(t')} \biggl( \sum_{t=1}^{T}{(x_{r}(t) - s(t))^2}}\biggr)\nonumber\\
&= -\sigma^{-2}\sum_{r=1}^{R}{(x_{r}(t') - s(t'))}.
\end{align}
Setting this equal to zero and solving for $s(t')$ we get
\begin{equation}
s(t') = \frac{1}{R} \sum_{1}^{R}{x_{r}(t')},
\end{equation}
which shows that if you believe that there is only one stereotypic
signal in the data then averaging the data over trials will yield
the optimal estimator of the source signal.

However, researchers are well aware that there are multiple
simultaneous signals, and that these signals vary from
trial-to-trial.  We have shown that both amplitude and latency
variability play a role in the variations of the signals emitted
by neural sources \cite{wilson1}.  This has led us to a new model
of the recorded signal from a set of neural sources
\begin{equation} \label{eq:single-sensor-model}
x_{r}(t) = \sum_{n=1}^{N}{\alpha_{nr} s_n(t - \tau_{nr}) +
\eta_r(t)},
\end{equation}
where $\alpha_{nr}$ describes the amplitude scale of the $n^{th}$
component during the $r^{th}$ trial, and $\tau_{nr}$ similarly
describes its latency shift forward or backward in time.  This
allows us to account for and to characterize amplitude changes and
response delays in the neural responses during the course of an
experiment or under different experimental conditions.  This model
assumes that each of the $N$ sources has a distinct stereotypic
waveshape. In our work we have found that by simply describing
these additional characteristics of the neural responses, we can
separate source signals that vary differentially from trial to
trial.  The algorithm that results from this model, and our
subsequent probability assignments, is called \emph{differentially
Variable Component Analysis} (dVCA) \cite{Knuth+etal:ICA01,
Truccolo+etal:2003, Knuth:inprep}.  To accommodate multiple
detectors, we simply modify the signal model accordingly
\begin{equation} \label{eq:multiple-sensor-model}
x_{mr}(t) = \sum_{n=1}^{N}{C_{mn} \alpha_{nr} s_n(t - \tau_{nr}) +
\eta_{mr}(t)},
\end{equation}
where $\mathbf{C}$ is the mixing matrix, or coupling matrix as we
call it since it describes the coupling between the sources and
the detectors.  With this new signal model in hand, we are already
making interesting new discoveries in our old data sets.

\section{Conclusion}

In this tutorial I have introduced the idea of \emph{informed
source separation}. My motivations here are those of a physical
scientist, where I have specific problems in need of accurate
solutions. In these cases, it is much more advantageous to begin
with the appropriate model, introduce the known prior information,
and derive an algorithm specifically engineered for the task.

Historically, while source separation had its beginnings in neural
networks and information theory
\cite{Jutten&Herault:1991,Pham&etal:1992,Cichocki&etal:1994,Bell&Sejnowski},
it was recognized early on that these results were related to the
maximum likelihood formalism
\cite{Cardoso,MacKay,Pearlmutter&Parra}. From this point, one is
easily led to the Bayesian methodology
\cite{Knuth:ME97,Roberts:1998,Knuth:ICA99,Mohammad-Djafari:1999,Rowe:1999}.
A distinct advantage of the Bayesian approach is that it breaks
the problem into three pieces: the signal model, the cost
function, and the search algorithm. The researcher begins by
choosing an appropriate signal model for the physical problem.
Once this model has been chosen, the researcher uses probability
theory to derive the posterior probability, which is the cost
function to be optimized. With a cost function in hand, a search
algorithm is employed to identify the optimal model parameter
values.  Each of these three pieces can be modified leading to
different algorithms that vary in applicability, accuracy and
efficiency.

Missing from this short tutorial is a discussion of the numerous
techniques and algorithms that can be used to search the parameter
space to identify solutions with high probabilities. I will
attempt to refer the reader to a variety of useful and important
techniques that have been presented in the literature. These
methods include: gradient ascent search
\cite{Bell&Sejnowski,Amari:1998}, iterative fixed point algorithms
\cite{Hyvarinen&Oja:1997,Knuth+etal:ICA01}, Markov chain Monte
Carlo (MCMC) \cite{Senecal&Amblard:2000,Fevotte&Godsill:2005},
sequential MCMC (also known as particle filters)
\cite{Everson&Roberts:2000,Larocque&Reilly&Ng:2002}, mean field
and ensemble methods
\cite{Lappalainen:1999,Miskin&MacKay:2000,Hojen-Sorensen&etal:2001,Valpola&Karhunen:2002},
variational Bayes
\cite{Lawrence&Bishop:2000,Choudry&Roberts:2001}, as well as
Bayesian techniques which utilize sparsity
\cite{Fevotte&Godsill:2005}. Last, as an aid to better
understanding Bayesian methods, I would recommend the following
introductory references \cite{Sivia:1996, Skilling:1998,
Dose:2002}.

I would also recommend that the reader seek out the other papers
presented in this special session to get a taste for the wide
array of methods and applications.  My hopes are that this
tutorial will inspire and enable readers to engineer algorithms
for their specific problems.



\end{document}